\documentclass{styles/svproc}
\usepackage{pgfpages}
\pgfpagesuselayout{resize to}[a4paper]
\usepackage{url}
\usepackage{graphicx}
\usepackage{amssymb}
\usepackage{amsmath}
\usepackage{multirow}
\usepackage{dblfloatfix} 
\usepackage{balance}
\usepackage{hyperref}
\usepackage{dsfont}
\usepackage{subcaption}

\makeatletter
\newcommand*\bigcdot{\mathpalette\bigcdot@{.5}}
\newcommand*\bigcdot@[2]{\mathbin{\vcenter{\hbox{\scalebox{#2}{$\m@th#1\bullet$}}}}}
\newcommand\norm[1]{\left\lVert#1\right\rVert}
\makeatother

\begin{document}
\mainmatter              % start of a contribution

\title{Model-free Visual Control for Continuum Robot Manipulators via Orientation Adaptation}
\titlerunning{Continuum Manipulator Orientation Adaptation}  % abbreviated title (for running head)
\author{Mrinal Verghese \and Florian Richter \and
Aaron Gunn\and Phil Weissbrod \and \\ Michael Yip}
\authorrunning{Verghese et al.} % abbreviated author list (for running head)
\institute{University of California, San Diego, La Jolla CA 92032, USA,\\
\email{\{mtverghe, frichter, ahgunn, pweissbrod, yip\}@ucsd.edu}
}

\maketitle              % typeset the title of the contribution

\begin{abstract}
We present an orientation adaptive controller to compensate for the effects of highly constrained environments on continuum manipulator actuation. A transformation matrix updated using optimal estimation techniques from optical flow measurements captured by the distal camera is composed with any Jacobian estimation or kinematic model to compensate for these effects. By utilizing domain knowledge to define the structure of this matrix, fewer parameters need to be estimated and a stable controller can be guaranteed. The algorithm is tested on a custom robotic catheter and convergence is shown both empirically and theoretically.

\keywords{Soft Robotics, Control}
\end{abstract}
%
% \begin{figure}[t]
%     \includegraphics[width=10cm]{figures/IntroConvergance.png}
%     \centering
%     \caption{Convergence of the instrument tip to a target point with and without our adaptive controller to compensate for unknown contact forces in heavily constrained environments. }
%     \label{fig:IntroConvergance}
% \end{figure}
\begin{figure}[t]
	\centering
	\begin{subfigure}{.65\textwidth}
		\centering
  		\includegraphics[width=1\linewidth]{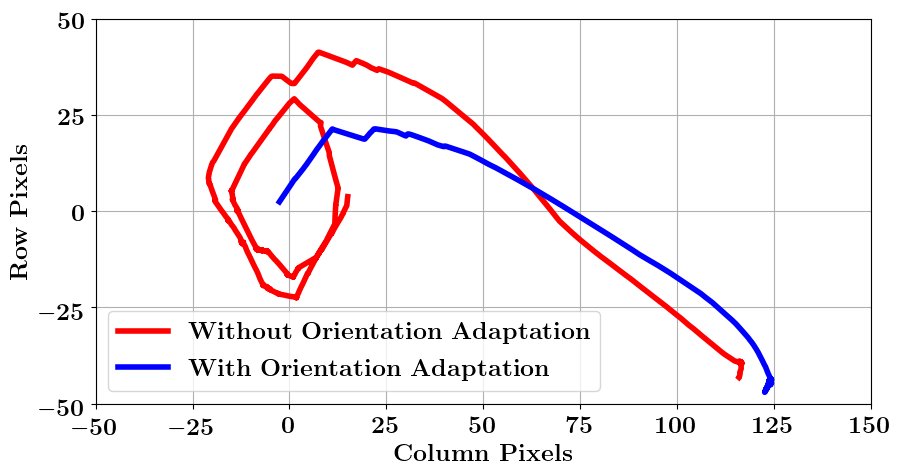}
	\end{subfigure}
	\hspace{1mm}
	\begin{subfigure}{.3\textwidth}
		\centering
  		\includegraphics[width=1\linewidth]{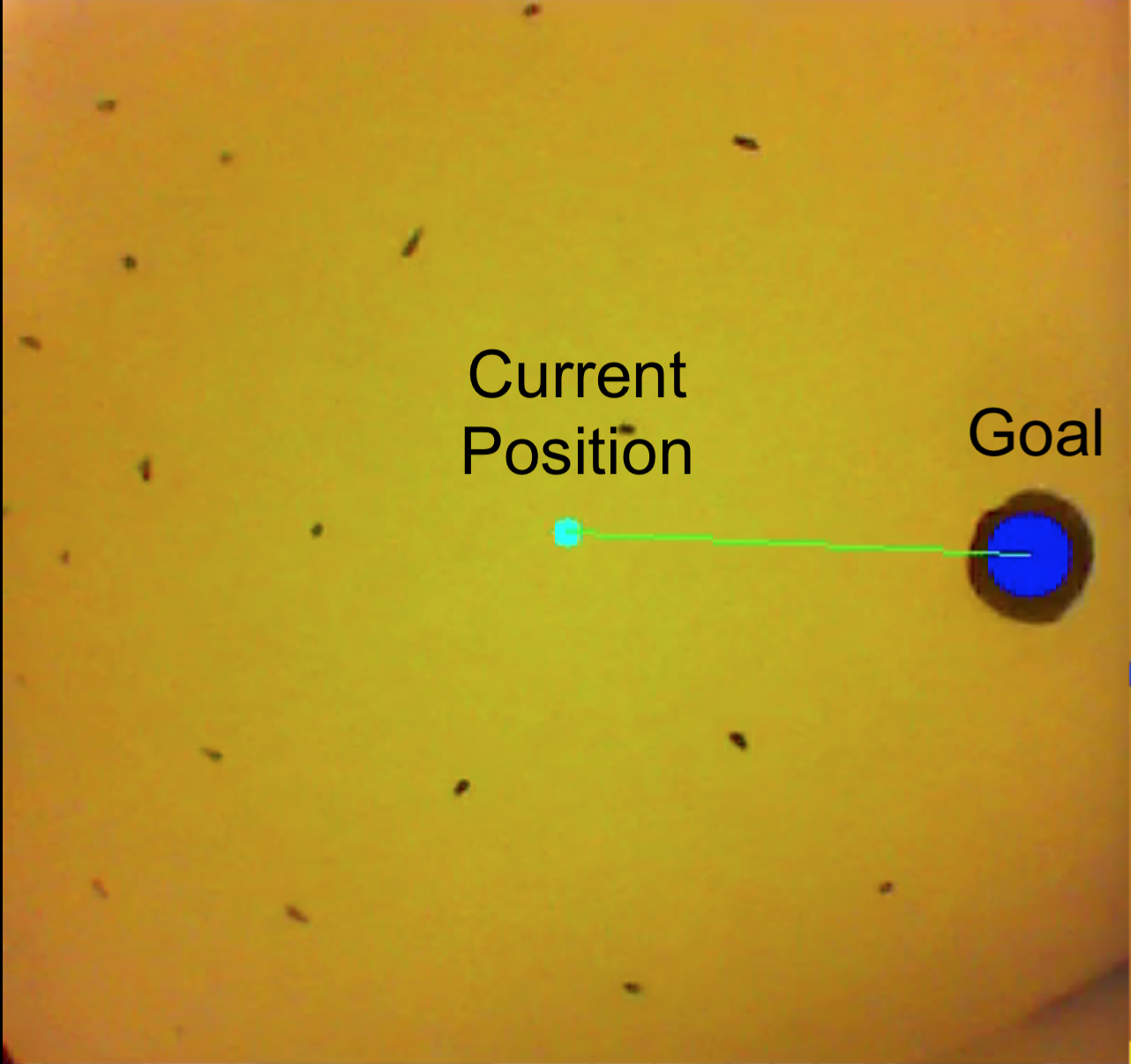}
  		\vspace{1mm}
	\end{subfigure}
	\caption{The plot on the left shows convergence of a target point (in pixel space) with and without our adaptive controller to compensate for unknown contact forces in a heavily constrained environment. The first person endoscopic camera is shown on the right viewing a test fixture and showing the  objective.}
	\label{fig:IntroConvergance}
\end{figure}

\section{Introduction}
Continuum manipulators have shown great promise for Minimally Invasive Surgery in the recent years \cite{J. Burgner-Kahrs,V. Vitiello}. Their ability to conform to natural anatomy allows greater access within the body without requiring incisions or trauma to the patient. This success has been evidenced by viable commercial products like the Monarch platform\texttrademark{} from Auris Surgical Robotics or the Ion Endoluminal System\texttrademark{} from Intuitive Surgical. Unfortunately, the soft structure of continuum robots bring significant control challenges along with their advantages. The robot has infinite degrees of freedom, and constrained space operation in human anatomy involves unknown contact forces and dynamics that mask measurements of the full robot configuration thus further complicating control \cite{R. S. Penning1,R. S. Penning2,A. Kapadia,D. Camarillo,V. K. Chitrakaran}. At worst, these contacts may result in unstable and dangerous behaviors \cite{M. Yip1}. %With these challenges in mind, we turn to adaptive control to achieve intuitive and predictable actuation of these instruments.%

\subsection{Background}
Multiple ideas have been tested against the constrained space control problem. The most generic strategy is to estimate the Jacobian matrix online \cite{M. Yip1,M. Yip2}, which was shown for setpoint / trajectory regulation as well as hybrid position-force control, and was extended by Wang \textit{et al.} via an adaptive visual servo controller \cite{H. Wang1,H. Wang2}. Efforts have also been made to sense the environment or the continuum manipulator configuration and utilize that information \cite{M. Yip3}.  Tully \textit{et al.} examined estimating the configuration of a segmented snake robot using an extended Kalman filter on the data from a distal electromagnetic tracker \cite{S. Tully}. Other learning methods have also been applied; Melingui \textit{et al.} implemented an adaptive control algorithm utilizing kernel-based learning for continuum manipulators in free space \cite{A. Melingui}, and Giorelli \textit{et al.} used a neural network to learn the kinematics of three-tendon actuators \cite{M. Giorelli}. As a whole, actuating continuum manipulators in free space or trivially constrained environments is generally well understood and can be done with moderate accuracy, whereas the problem of reliable control in heavily constrained space such as in many surgical tasks is challenging and open ended.

\subsection{Contributions}
  Our previous work examined a completely model-less control framework to solve this problem by estimating the Jacobian online from a distal camera and tendon tension sensors \cite{M. Yip1,M. Yip2,M. Yip3}. However, without structure to the estimated Jacobian, there is a risk of drift in the matrix or potential singularities. We look to build on our previous work by utilizing domain knowledge about the control problem to derive more structured transformations based on fewer parameters that can be estimated from relatively minimal observations. These parameters are then optimized online to ensure convergence and stability in workspace tasks. We apply this controller on a real robotic catheter to verify our claims and demonstrate the advantages of our methods.

\section{Methods}

When operating tendon based robotic catheters, any actuation is defined by a linear combination of changes in the lengths of multiple tendons. This actuation can be reasonably modeled in free space by constant curvature models \cite{R. J. Webster}. In constrained space operation, however, unknown contact forces can change the distributed friction along tendon lumens resulting in unpredictable configurations (see Fig \ref{fig:IntroConvergance}). For control in the distal camera frame, this results in a control mismatch between control inputs and actual actuation. This poses problems both for human operators who expect consistent control while navigating the body, and autonomous algorithms such as visual servoing, whose feedback controllers rely on accurate models for convergence to setpoints and trajectories.

\subsection{Problem Formulation}
\begin{figure}[b]
    \includegraphics[width=12cm]{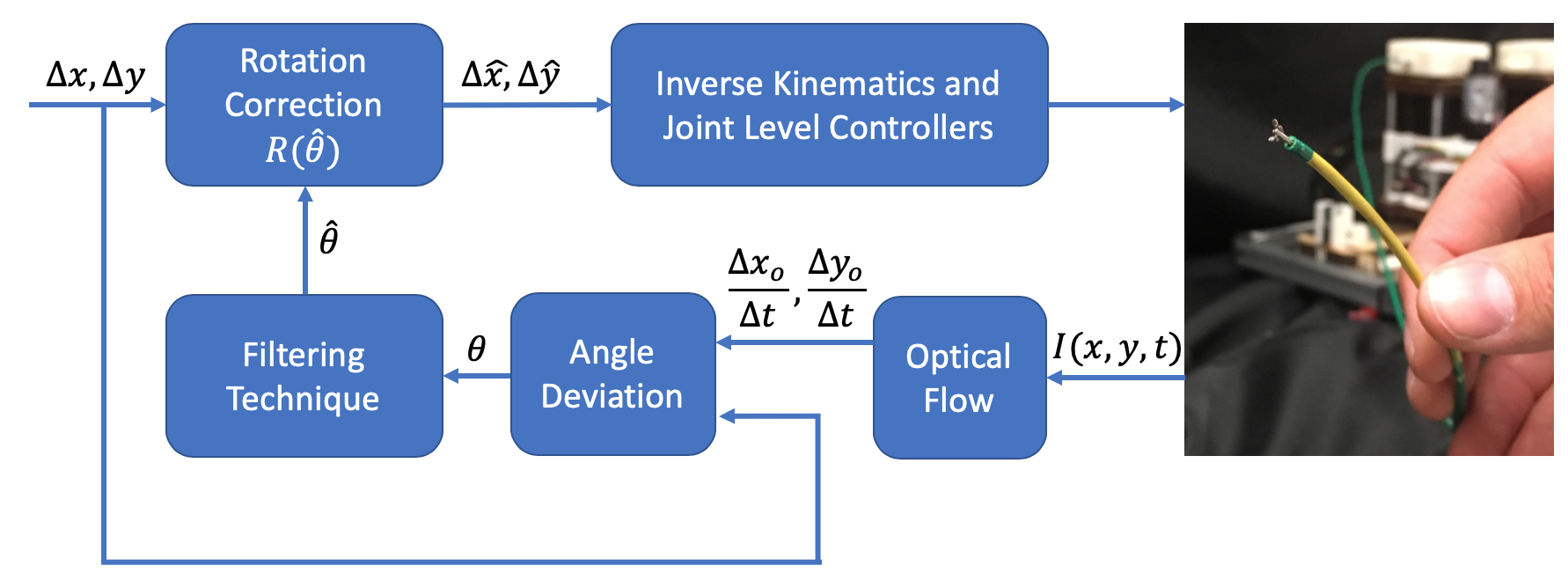}
    \centering
    \caption{Control flow chart that shows the process for model-free learning applied to Jacobian rotation adaptation using visual feedback.}
    \label{fig:flowchart}
\end{figure}
We consider the problem of model-less control \cite{M. Yip1} for endoscopes with video feedback, and with  setpoints and trajectories defined in pixel space. Actuation is considered in the camera frame where the $z$ axis is the depth, and the $x, y$ axes correspond to the pixel columns and rows respectively. Typically the $z$ axis is controlled by a separate linear insertion joint, and actuation in this direction is not considered in this work. The Jacobian, $\mathbf{J}$, evaluated at configuration $\mathbf{q}$, relates the actuator velocities to the camera center movement in pixel space:
\begin{equation*}
    \begin{bmatrix} \dot{x} & \dot{y} \end{bmatrix}^{\top} \approx \mathbf{J} \begin{bmatrix} \dot{q_1} & \dot{q_2} & \dots & \dot{q_n} \end{bmatrix}^{\top}
\end{equation*}
where $\mathbf{\dot{q}} = \begin{bmatrix} \dot{q_1} & \dot{q_2} & \dots & \dot{q_n} \end{bmatrix}^{\top}$ are the joint velocities. The pseudo-inverse of the Jacobian, $\mathbf{J}^\dagger$, is used to convert control inputs from the camera frame to changes in joint angles. In constrained conditions, and for that matter, in situations where the kinematic model of the robot is inexact, the model-based Jacobian is inaccurate. Formally we describe this problem as:
\begin{equation}
    \mathbf{J}_* \mathbf{J}^\dagger \begin{bmatrix} \dot{x} & \dot{y} \end{bmatrix}^{\top} \neq \begin{bmatrix} \dot{x} & \dot{y} \end{bmatrix}^{\top}
\end{equation}
where $\mathbf{J}_*$ is the real Jacobian, and the left-hand-side and right-hand-side of the equations represent camera motion vectors in the image frame that are dissimilar. 

In this work, corrections are applied by estimating an approximate rotation matrix that, when applied to the Jacobian model $\mathbf{J}$, where the rotation is found by minimizing the error between these two vectors in a robust manner (i.e. accounting for measurement and process noise). This process is shown in Fig. \ref{fig:flowchart}.
%What were we trying to say here?
Note that these corrections can be combined with other methods of estimation or corrections of the Jacobian, as correcting for rotation alone does not necessarily converge to zero error due to time and history-dependent effects such as viscoelasticity, creep, and hysteresis. Formally, this correction when converged should result in:

\begin{equation}
    \frac{ \mathbf{J}_* \mathbf{J}^\dagger \mathbf{R}(\theta) \begin{bmatrix} \dot{x} & \dot{y} \end{bmatrix}^{\top} }{\norm{\mathbf{J}_* \mathbf{J}^\dagger \mathbf{R}(\theta) \begin{bmatrix} \dot{x} & \dot{y} \end{bmatrix}^{\top}}} =  \frac{ \begin{bmatrix} \dot{x} & \dot{y} \end{bmatrix}^{\top} } {\norm{\begin{bmatrix} \dot{x} & \dot{y} \end{bmatrix}^{\top}}}
    \label{equation:goal}
\end{equation}
where $\mathbf{R}(\theta)$ is the rotation matrix correcting the Jacobian is estimated from measurements captured by the endoscopic camera, and $\norm{\cdot}$ represents a vector l2-norm.

For an intuitive derivation of the correction, let $\mathbf{R} \in \mathbb{R}^{2 \times 2}$ be a a general linear correction on the control input. It can be decomposed using singular-value decomposition (SVD):
\begin{equation}
    \mathbf{R} = \mathbf{U} \begin{bmatrix} \sigma_1 & 0 \\ 0 & \sigma_2 \end{bmatrix} \mathbf{V}^{\top}
\end{equation}
where $\mathbf{U}$ and $\mathbf{V}$ are unitary matrices and $\sigma_1$ and $\sigma_2$ are the singular values. 

If $\sigma_1 \neq \sigma_2$, then the $\mathbf{R}$ describes a shear, which is a case left for future work. For this work, it is a assumed that $\sigma_1 = \sigma_2$.

This implies that $\mathbf{R}$ can be written as a product of a scalar and two unitary matrices, which simplifies to a scalar and a single unitary matrix. The scalar can be seen as the required compensation on the magnitude of the control input to overcome any loss of energy in the system due to frictional and viscoelastic losses from actuation and interaction with the environment. This is near-impossible to estimate due to the non-linear and temporal effects as well as the limited sensing information of where contact occurs within the environment. For purposes of estimation, this scalar is set to 1 and only the unitary matrix is left to estimate. The unitary matrix can be rewritten as a rotation matrix:
\begin{equation}
    \mathbf{R}(\theta) = \begin{bmatrix} \text{cos}(\theta) & -\text{sin}(\theta) \\ \text{sin}(\theta) & \text{cos}(\theta) \end{bmatrix}
\end{equation}
which only dependent on a single parameter $\theta$.

\subsection{Measurement}

The image data at point $x,y$ and time $t$ is written as $\mathbf{I}(x,y,t)$. Assuming a small motion between frames and time step, a first order approximation for the image data can be written as:
\begin{equation}
    \mathbf{I}(x + \Delta x_o, y + \Delta y_o, t + \Delta t) = \mathbf{I}(x,y,t) + \frac{\partial \mathbf{I}}{\partial x}\Delta x_o + \frac{\partial \mathbf{I}}{\partial y}\Delta y_o + \frac{\partial \mathbf{I}}{\partial t}\Delta t
    \label{equation:first_order_image}
\end{equation}
where $\Delta x_o, \Delta y_o$ is the observed motions from the endoscopic camera. Using the previous notation, the observed motion can be defined by:
\begin{equation}
    \mathbf{J}_* \mathbf{J}^\dagger \begin{bmatrix} \Delta x & \Delta y \end{bmatrix}^{\top} = \begin{bmatrix} \Delta x_o &  \Delta y_o \end{bmatrix}^{\top}
    \label{equation:measurement}
\end{equation}
where the changes in the discretized control are also assumed to be small. Combining (\ref{equation:measurement}) and (\ref{equation:goal}) with $\Delta x_o, \Delta y_o$ as $\dot{x}, \dot{y}$ results in:
\begin{equation}
    \frac{ \mathbf{R}(\theta) \begin{bmatrix} \Delta x_o & \Delta y_o \end{bmatrix}^{\top} }{\norm{\mathbf{R}(\theta) \begin{bmatrix} \Delta x_o & \Delta y_o \end{bmatrix}^{\top}}} = \frac{\begin{bmatrix} \Delta x & \Delta y \end{bmatrix}^{\top}}{\norm{\begin{bmatrix} \Delta x & \Delta y \end{bmatrix}^{\top}}} \text{ }.
\end{equation}
Note that the Jacobian, $\mathbf{J}_* \mathbf{J}^\dagger \in \mathbf{R}^{2 \times 2}$, is assumed to be a full rank to get this expression. The cases where this does not hold is if the control input does not overcome the losses of energy in the system due to frictional and viscoelastic losses, which is not covered in this work, or if a end-effector collision occurs (which necessitates some additional measurements of contact, as demonstrated in \cite{M. Yip2}). By simply measuring the angle between the intended control, $\Delta x, \Delta y$, and the observed motion $\Delta x_o, \Delta y_o$ yields the angle $\theta$, required for the correction.

Following the brightness constancy constraint, which says projection of the same point results in the same image data at every frame, the first order approximation for the image data from (\ref{equation:first_order_image}) can be simplified to:
\begin{align}
    &\frac{\partial \mathbf{I}}{\partial x}\Delta x_o + \frac{\partial \mathbf{I}}{\partial y}\Delta y_o + \frac{\partial \mathbf{I}}{\partial t}\Delta t = 0\\
    &\frac{\partial \mathbf{I}}{\partial x}v_x + \frac{\partial \mathbf{I}}{\partial y}v_y = - \frac{\partial \mathbf{I}}{\partial t}
\end{align}
where $\mathbf{v} = \begin{bmatrix}v_x&  v_y \end{bmatrix}^{\top}$ is the observed optical flow. Therefore, optical flow is directly proportional to the observed motion of continuum manipulator in the camera frame. Optical flow can be measured in a variety of ways; the Lucas-Kanade method utilizing Shi-Tomasi corner detection is used for this work \cite{Optical Flow,Good Features to Track}.

The angle between the observed motion and intended motion can be found by comparing the optical flow result and the control input $\Delta x, \Delta y$ and is the correction needed. The angle between the two vectors is simply computed:
\begin{equation}
    \theta = \text{cos}^{-1}\Bigg( \frac{  \begin{bmatrix} \Delta x & \Delta y \end{bmatrix}^\top \bigcdot \mathbf{v}}{ \norm{\begin{bmatrix} \Delta x & \Delta y \end{bmatrix}^\top}\text{ } \norm{\mathbf{v}} } \Bigg)
\end{equation}
and $\theta$ generates the linear correction $\mathbf{R}(\theta)$.

\subsection{Estimation}
A measurement of $\theta(t)$ can be taken at every image pair $\mathbf{I}(\cdot, \cdot, t)$ and  $\mathbf{I}(\cdot, \cdot, t-1)$ and is generated from optical flow which is very noisy. The Kalman Fitler can be used to reduce noise and have a more accurate measurement. To formulate this as a filtering problem, let $\hat{\theta}(t)$ be the filtered estimate of the correction with variance $\sigma^2_{\theta}(t)$. The motion model is assumed to have Gaussian noise, so the equation is simply:
\begin{equation}
    \hat{\theta}(t) = \hat{\theta}(t-1) + w
\end{equation}
where $w \sim \mathcal{N}(0, \sigma_w^2)$. Similarly, the measurement model is assumed to have Gaussian noise resulting:
\begin{equation}
    \theta(t) = \hat{\theta}(t) + v
\end{equation}
where $v \sim \mathcal{N}(0, \sigma_v^2)$. Using the Kalman Filter update, the estimate evolves:
\begin{align}
    &k(t) = \frac{\sigma^2_{\theta}(t-1) + \sigma_w^2}{\sigma^2_{\theta}(t-1) + \sigma_w^2 + \sigma_v^2} \\
    &\hat{\theta}(t) = k(t) \hat{\theta}(t-1) + \big(1 - k(t)\big)\theta(t) \\
    &\sigma^2_{\theta}(t) = ( 1 - k(t)) (\sigma^2_{\theta}(t-1) + \sigma_w^2)
\end{align}
where $k(t)$ is the Kalman gain. This formulation fits the requirements for convergence on the Kalman gain, $k(t)$ \cite{Kalman Filter Converge}, so the filter will converge to an Infinite Impulse Reponse (IIR) filter:
\begin{equation}
    \hat{\theta}(t) = \alpha \hat{\theta}(t-1) + \big(1 - \alpha \big)\theta(t) 
\end{equation}
based on a single parameter $\alpha$. It is easier to tune the single parameter $\alpha$, rather than all the parameters required for the Kalman Filter: $\sigma_{\theta}(0)$, $\sigma_w$, and $\sigma_v$, so the converged IIR filter is used for estimation. In order to account for the angle wrap around, the final filter used for estimation is:
\begin{equation}
    \hat{\theta}(t) = \text{mod}\Big(\alpha \hat{\theta}(t-1) + \big(1 - \alpha \big)\theta(t) + \pi, 2\pi\Big) - \pi
\end{equation}
to bound the estimate between $[-\pi, \pi)$. 

Additionally, a threshold is placed on the magnitude from measured optical flow, $||\mathbf{v}||$, in order to do an update. This avoids the measurements where the control input does not overcome the losses of energy in the system, which, as previously described, is when $\mathbf{J_*J}^\dagger$ is not full rank. 

\subsection{Lyapunov Stability}
We show under ideal conditions (no hold effects, delay, etc.) the rotation adaptation controller is stable. Let the error be $\mathbf{e} = \begin{bmatrix}  x_d & y_d \end{bmatrix}^\top - \begin{bmatrix}  x & y \end{bmatrix}^\top$, where $x_d, y_d$ is some desired position and $x,y$ be current position, and the candidate Lyapunov function is $V = \frac{1}{2}\mathbf{e}^\top \mathbf{e}$. The resulting derivative of this is:
\begin{equation}
    \dot{V} = \mathbf{e}^\top \begin{bmatrix} \dot{x_d} \\ \dot{y_d} \end{bmatrix} - \mathbf{e}^\top \begin{bmatrix}  \dot{x} \\ \dot{y} \end{bmatrix} \text{ .}
    \label{equation:lyapunov}
\end{equation} 
In standard position regulation the desired positions would not change, so $\dot{x_d}, \dot{y_d}$ are both 0, and the error $\mathbf{e}$, would be inputted into the adaptive controller system described in Fig. \ref{fig:flowchart}. Therefore, the instantaneous change in output would be:
\begin{equation}
    \begin{bmatrix}  \dot{x} \\ \dot{y} \end{bmatrix} = \mathbf{J}_* \mathbf{J}^\dagger \mathbf{R} (\theta) \mathbf{e} \text{ .}
\end{equation} 
Substituting this into (\ref{equation:lyapunov}), the resulting derivative of the Lyapunov candidate function is:
\begin{equation}
    \dot{V} = - \mathbf{e}^\top \mathbf{J}_* \mathbf{J}^\dagger \mathbf{R} (\theta) \mathbf{e} \text{ }.
\end{equation} 
Through the adaptive controller, the values of $\theta$ are set such that (\ref{equation:goal}) is satisfied. Therefore the derivative of the Lyapunov candidate function chosen here is: 
\begin{equation}
\dot{V} = - \frac{|| \mathbf{J}_* \mathbf{J}^\dagger \mathbf{R} (\theta) \mathbf{e} || }{||\mathbf{e}||} 
\mathbf{e}^\top \mathbf{e}
\end{equation} which clearly is always negative except for when the error reaches 0 or when $\mathbf{J_*J}^\dagger$ is not full rank. Therefore, the system is considered asymptotically stable assuming the control input overcomes the losses of energy in the system, which as stated previously is when $\mathbf{J_*J}^\dagger$ is not full rank. 

\section{Experiments}

\begin{figure}[b]
	\centering
	\begin{subfigure}{.3\textwidth}
		\centering
  		\includegraphics[width=1\linewidth]{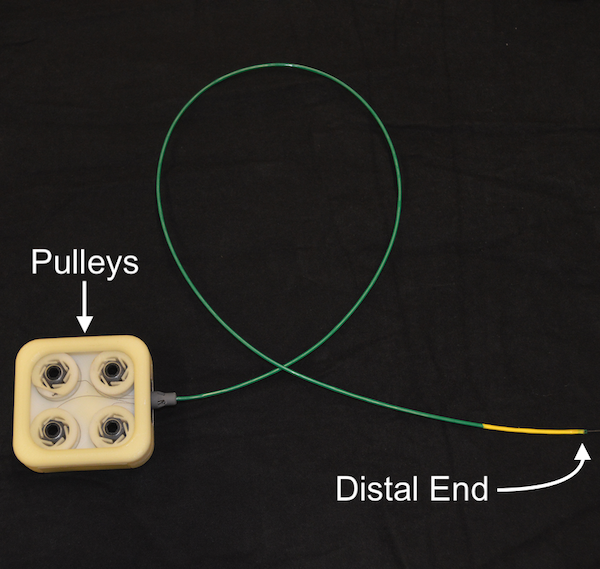}
	\end{subfigure}
	\hspace{1mm}
	\begin{subfigure}{.3\textwidth}
		\centering
  		\includegraphics[width=1\linewidth]{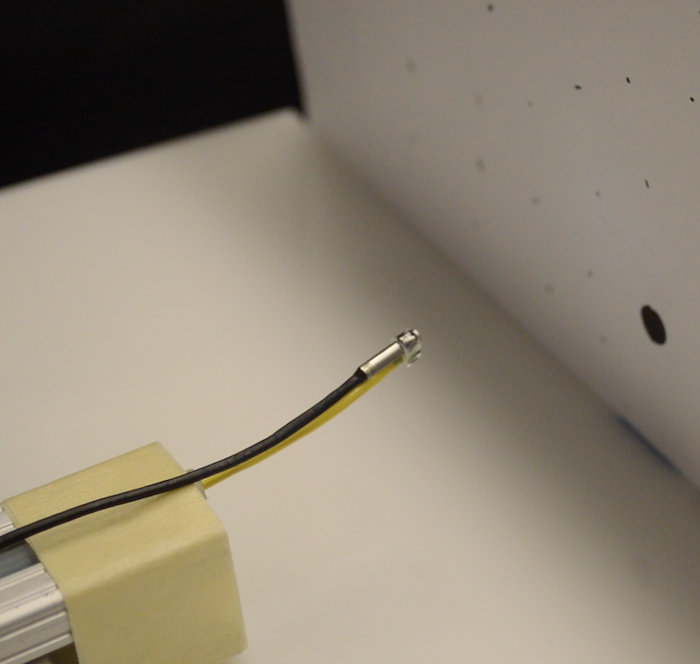}
	\end{subfigure}
	\hspace{1mm}
	\begin{subfigure}{.3\textwidth}
		\centering
  		\includegraphics[width=1\linewidth]{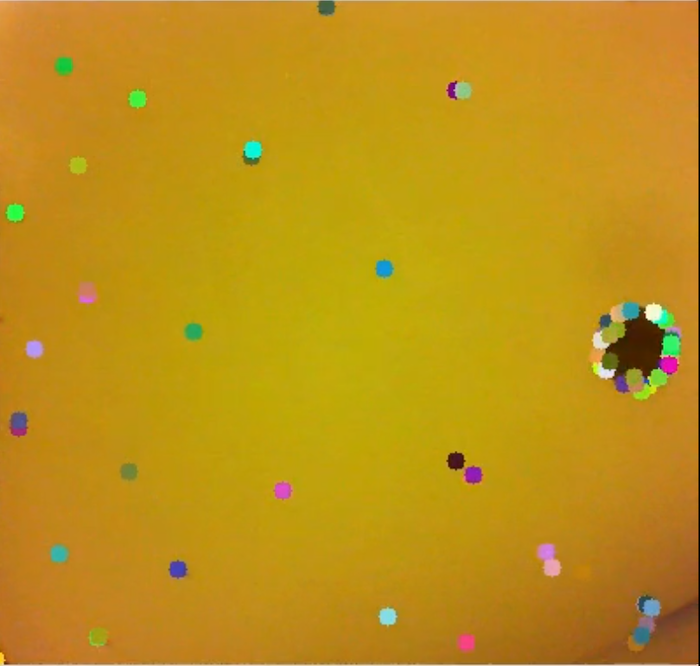}
	\end{subfigure}
	\caption{Full view of robotic catheter (left), view of the experimental setup at the distal end (middle), and camera view highlighting the optical flow measurement (right). Features to track for the optical flow were manually added in the environment for this experiment to ensure consistency between tests.}
	\label{fig:endoscope_view}
\end{figure}

\begin{figure}[t]
	\centering
	\begin{subfigure}{0.3\textwidth}
		\centering
  		\includegraphics[width=1\linewidth]{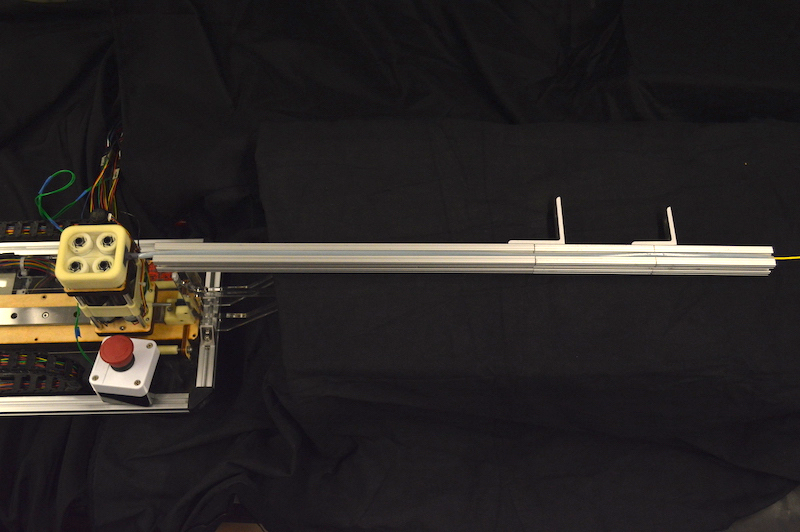}
	\end{subfigure}
	\hspace{1mm}
	\begin{subfigure}{.3\textwidth}
		\centering
  		\includegraphics[width=1\linewidth]{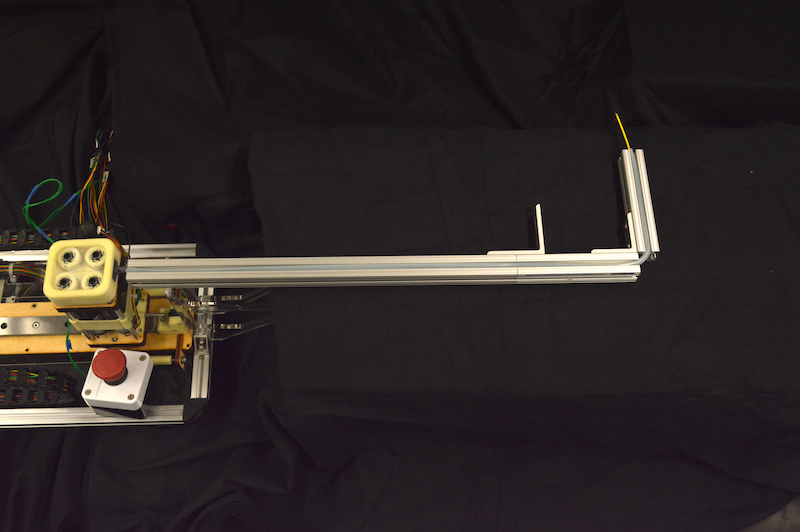}
	\end{subfigure}
	\hspace{1mm}
	\begin{subfigure}{.3\textwidth}
		\centering
  		\includegraphics[width=1\linewidth]{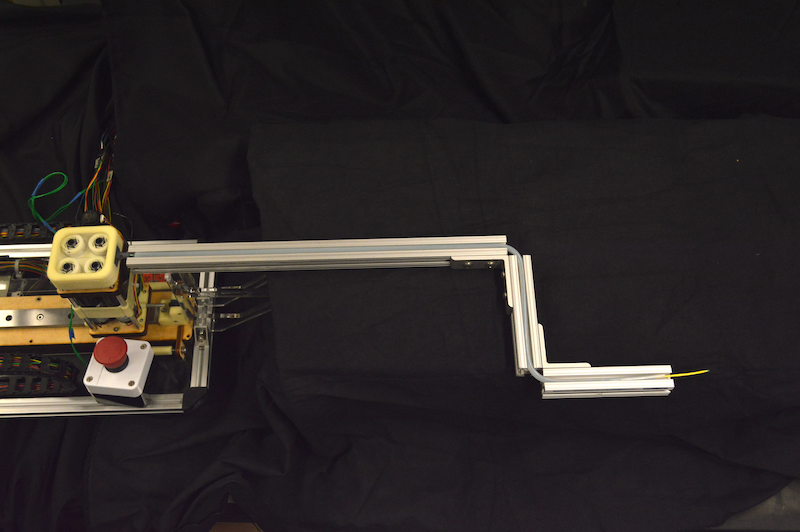}
	\end{subfigure}
	\caption{The three environments that the custom made robotic catheter is tested on. From left to right, the environments are named no bend, one bend, and two bend.}
	\label{fig:test_environments}
\end{figure}

% Add labelled photo of robot
% Experiment set up, color correct and more even spacing!!

To run the experiments we designed and built a flexible 2.2 mm diameter, meter-long robotic catheter. The robotic catheter is a continuum robot with a flexible backbone comprising five inner channels, one central backbone and four equally spaced (90$^{\circ}$ separation) around the perimeter. Stainless steel actuation wires/tendons routed through the four radially spaced channels and terminated at the distal end of the catheter provide deflections at the catheter tip. Tendons are terminated on the pulleys in the proximal end, and each pulley is controlled by a separate DC motor. Two tendons, 180$^\circ$ apart represent one antagonistic pair that controls deflection approximately on one plane. Insertion of the catheter is controlled by a separate DC motor that moves the entire assembly on a single linear rail.

All motors are driven by a custom FPGA motor controller system running individual PD loops on each motor \cite{D. Schreiber}. Control was completed utilizing inverse kinematics from the camera frame to the actuators from constant curvature models \cite{R. J. Webster}.  The controller interfaces over ethernet to a desktop running our algorithm in Python. Finally, an endoscopic camera with a total diameter of 2.5 mm with a resolution of 380x400 pixels, a framerate of 30 fps and an integrated LED light ring is attached to the distal end and connected to the desktop through USB. All image processing is done through Python's OpenCV library.

% \begin{figure}[t]
%     \includegraphics[width=5cm]{figures/CatheterContrsuction.PNG}
%     \centering
%     \caption{Cross-sectional layout of robotic catheter construction}
%     \label{fig:CatheterContruction}
% \end{figure}

To test the algorithm in constrained environments, three tortuous paths were made from nylon tubing and metal frames. The environments have zero, one, and two bends and are shown in Fig. \ref{fig:test_environments}. The catheter is passed through these environments and then placed in front of an optical marker that was off center in the endoscopic camera's view, as shown in Fig \ref{fig:IntroConvergance}. A simple proportional-controller is used to actuate the catheter and attempt to align the center of the camera with the optical marker. This experimental set up is shown in Fig. \ref{fig:endoscope_view}. Each trial ran for roughly 70 seconds or until it either converged or clearly diverged. The pixel position of the marker and the current value of $\hat{\theta}$ were recorded. When operating a steerable catheter, the operator wants to center a target in the camera frame for insertion, thus pixel distance from the center of the camera feed to the target is an appropriate evaluation criteria. Finally, for each environment the trials were repeated with $\alpha = 1$ (no correction), and  $\alpha = .95, .75, .50$.

\section{Results}

\begin{figure}
	\centering
	\begin{subfigure}{.85\textwidth}
		\centering
  		\includegraphics[width=1\linewidth]{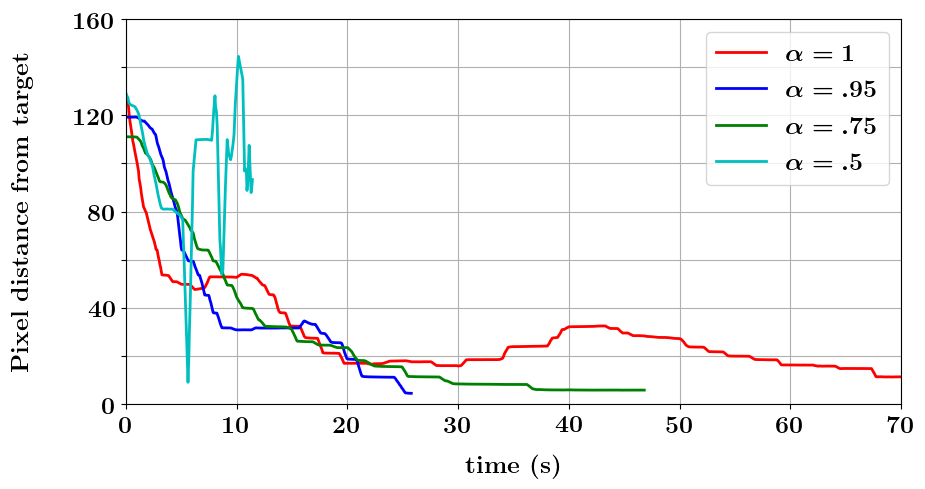}
	\end{subfigure}
    \vspace{1mm}
	\begin{subfigure}{.85\textwidth}
		\centering
  		\includegraphics[width=1\linewidth]{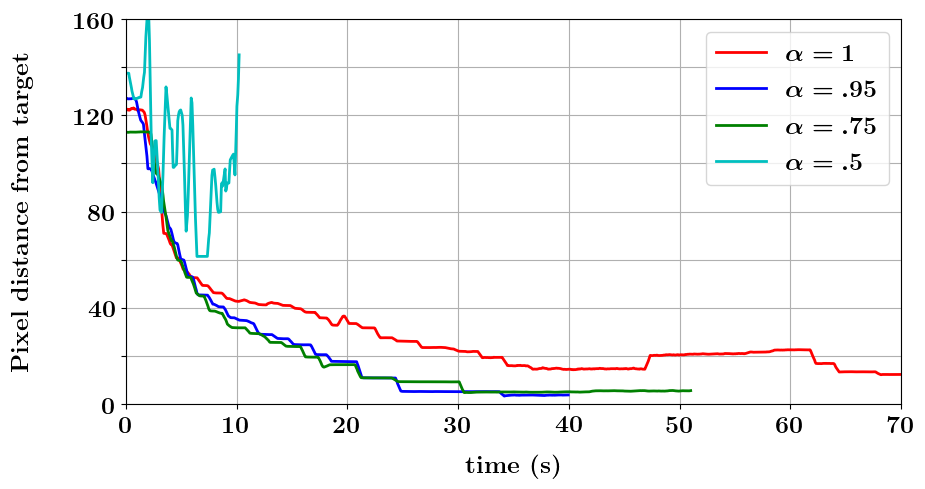}
	\end{subfigure}
	\vspace{1mm}
	\begin{subfigure}{.85\textwidth}
		\centering
  		\includegraphics[width=1\linewidth]{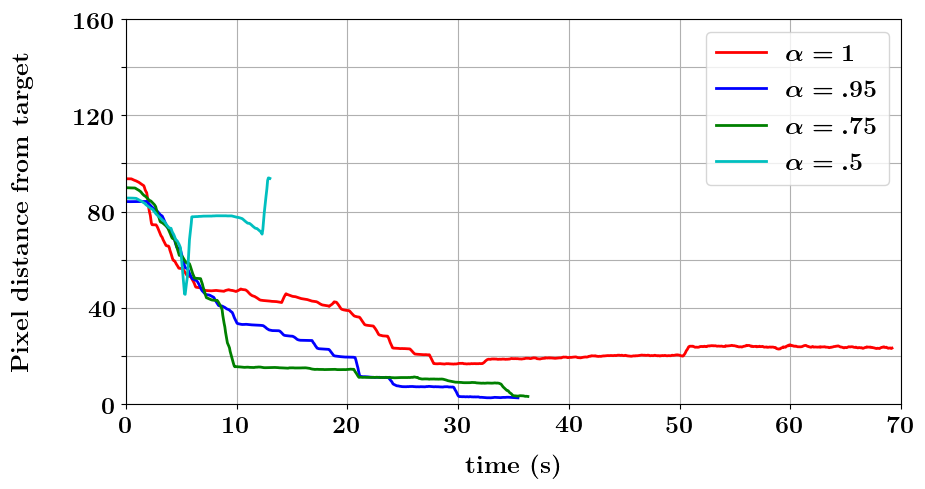}
	\end{subfigure}
	\caption{Pixel distance to target over time for three environments using different values of $\alpha$. From top to bottom the environments are: no bend, one bend, and two bends.}
	\label{fig:results_dist}
\end{figure}
The pixel distance to the target over time for all the experiments are shown in Fig. \ref{fig:results_dist}. The full trajectory of the pixel position of the goal for $\alpha = 1, 0.95$ for the one bend environment is shown in Fig. \ref{fig:IntroConvergance}. This visualization highlights the need for orientation adaptation when dealing with the more tortuous environments. The results for the robotic catheter without correction (when $\alpha = 1$) reaffirm the problem. Without any adaptation, the instrument has difficulty converging in increasingly constrained environments, including the no bend environment. The trajectory visualization in Fig. \ref{fig:IntroConvergance} highlights that when not using orientation adaptation, orbiting around target can occur due to the mismatch in observed and expected actuation and the the instrument may never reach said target. The same figure also shows the rapid convergence with the proposed orientation adaptation algorithm. In all tested environments, we were able to see the controller rapidly converge for $\alpha = .95, .75$. With $\alpha = .5$, the optical flow measurements were too noisy preventing $\hat{\theta}$ from converging and leading to instability in the controller.

\section{Discussions}

Results from our evaluation of adaptation of the Jacobian via a online rotation estimation show the advantages in accuracy and speed of rotation adaptation given different filtering choices.
Examining the plots for $\alpha = .95, .75$ shows the trade off between the two values. $\alpha = .75$ converges slightly faster in more constrained environments, at the expense of slightly nosier values for theta and and a less smooth trajectory. $\alpha = .95$ yielded a more reliable convergence across all our environments at the expense of taking longer to reach $\hat{\theta}$. Generally this parameter can be tuned based on the quality of the tracked features in the environment and the noise in optical flow.

\section{Conclusions and Future Work}
Our results showed that estimating only one parameter online still yielded fairly rapid convergence to an accurately mapped Jacobian matrix that would be needed for adaptive and stable control. Such adaptation has significant benefit to both autonomous tasks and obvious benefits to human teleoperators that would observe a correction such that steering commands match the camera directions exactly. In addition, as this single parameter is directly observable, the correction is not prone to drift or artificial singularities. While many completely model free approaches could also converge and generalize very well, integrating domain knowledge about the nature of the changes to the Jacobian helped build a more stable and efficient controller. Furthermore, utilizing a simple linear transform and its SVD decomposition presented in the problem formulation gives good intuition on additional parameters that can be estimated and what they physically represent. A tradeoff is made between the potential model capacity of purely model-free estimation and statistical robustness and stability of structured matrices defined by fewer parameters. This is in part due to the better physical understanding of the perturbations being made to continuum manipulator control to compensate for interactions with constrained environments. Ultimately this stability is essential to seeing this type of control realized in critical surgical applications.

%This work shows the effectiveness of single parameter estimation. Future work would be to explore other parameters such as shear and scaling to obtain a better linear transformation on the control input.


\end{document}